\documentclass[3p,11pt]{elsarticle}
\usepackage{graphicx}
\usepackage[dvipsnames]{xcolor}
\usepackage{amsmath}
\usepackage{amssymb}
\usepackage[ruled]{algorithm2e}
\usepackage{enumitem}
\usepackage[hidelinks]{hyperref}
\usepackage{booktabs}
\setlist{nolistsep}

\makeatletter
\def\ps@pprintTitle{%
  \let\@oddhead\@empty
  \let\@evenhead\@empty
  \let\@oddfoot\@empty
  \let\@evenfoot\@oddfoot}
\makeatother

\begin{document}

\begin{frontmatter}

\title{Mitigating Clever Hans Strategies in Image Classifiers through Generating Counterexamples}

\author[tub,bifold]{Sidney Bender}
\author[tub]{Ole Delzer}
\author[basf]{Jan Herrmann}
\author[basf]{Heike Antje Marxfeld}
\author[tub,bifold,korea,mpi]{Klaus-Robert Müller}
\author[bifold,charite]{Grégoire Montavon\corref{cor1}}
\ead{gregoire.montavon@charite.de}

\cortext[cor1]{Corresponding author}

\address[tub]{Machine Learning Group, Technische Universit\"at Berlin, 10587 Berlin, Germany}
\address[basf]{BASF SE, 67056 Ludwigshafen am Rhein, Germany}
\address[bifold]{BIFOLD\;--\;Berlin Institute for the Foundations of Learning and Data, 10587 Berlin, Germany}
\address[korea]{Department of Artificial Intelligence, Korea University, 02841 Seoul, Korea}
\address[mpi]{Max-Planck Institute for Informatics, 66123 Saarbr\"ucken, Germany}
\address[charite]{Charit\'e\;--\;Universit\"atsmedizin Berlin, 10115 Berlin, Germany}

\begin{abstract}
Deep learning models remain vulnerable to \emph{spurious correlations}, leading to so-called \emph{Clever Hans} predictors that undermine robustness even in large-scale foundation and self-supervised models. Group distributional robustness methods, such as Deep Feature Reweighting (DFR) rely on explicit group labels to upweight underrepresented subgroups, but face key limitations: (1) group labels are often unavailable, (2) low within-group sample sizes hinder coverage of the subgroup distribution, and (3) performance degrades sharply when multiple spurious correlations fragment the data into even smaller groups.
We propose Counterfactual Knowledge Distillation (CFKD), a framework that sidesteps these issues by generating diverse counterfactuals, enabling a human annotator to efficiently explore and correct the model's decision boundaries through a knowledge distillation step. Unlike DFR, our method not only reweights the undersampled groups, but it also enriches them with new data points. Our method does not require any confounder labels, achieves effective scaling to multiple confounders, and yields balanced generalization across groups.
We demonstrate CFKD’s efficacy across five datasets, spanning synthetic tasks to an industrial application, with particularly strong gains in low-data regimes with pronounced spurious correlations. Additionally, we provide an ablation study on the effect of the chosen counterfactual explainer and teacher model, highlighting their impact on robustness.
\end{abstract}

\end{frontmatter}

\section{Introduction}

Deep learning has achieved remarkable progress in recent years, delivering state-of-the-art performance across a wide range of domains, including computer vision, natural language processing, and biomedical applications. However, despite these advancements, models frequently rely on \emph{spurious features}—also known as \emph{confounders}—which can give rise to so-called \emph{Clever Hans} (CH) predictors \cite{pfungst1911clever,lapuschkin2019unmasking}. Such models may fit training data well and achieve high validation accuracy, yet fail catastrophically when deployed under realistic conditions. These failures are particularly concerning in safety-critical and regulated domains such as medical imaging, autonomous driving, and industrial quality control, where robustness and trustworthiness are non-negotiable requirements. Reflecting these concerns, the Organisation for Economic Co-operation and Development (OECD) has developed AI principles—the first intergovernmental standard on AI~\cite{oecd2024ai}—which emphasize transparency and accountability, and are, for example, interpreted in the EU as requiring AI actors to provide information on how their systems make decisions. Such transparency requirements are especially needed in the process of deploying a new AI model, to verify that the AI model's decision making is robust and does not rely on spurious correlations.

\medskip

A growing body of work has revealed that the reliance on spurious correlations is not merely a problem of small-scale or narrowly trained models. Even modern `foundation models' (FMs)---trained on vast and heterogeneous datasets with billions of parameters---remain vulnerable to learning non-causal, non-domain-relevant features~\cite{kauffmann2025explainable}. Recent evidence shows that CH effects are widespread: models often capture dataset-specific shortcuts that lead to brittle representations, with significant performance losses under even mild distribution shifts or subgroup reweightings. The biomedical domain comprises particularly critical instances of this general problem. Here, the stakes are uniquely high: studies on pathology FMs have demonstrated that such models can systematically encode non-biological technical artifacts---such as scanner hardware differences or laboratory-specific preparation procedures---which in turn severely compromise downstream diagnostic accuracy \cite{komen2025towards}. Taken together, these findings highlight that neither scaling up nor adopting self-supervised training paradigms inherently solves the problem of spurious correlations.

\medskip

Existing attempts to mitigate spurious correlations fall into three largely independent research directions. The first comprises generic data augmentation techniques (e.g.\ color jittering, mixup~\cite{zhang2017mixup}, geometric transformations, random augmentation with a diffusion model~\cite{trabucco2023effective}), which aim to break correlations between task labels and nuisance variables by enriching the training distribution. The second line of work (e.g.\ GroupDRO~\cite{sagawa2019distributionally}, JTT~\cite{liu2021just}, DFR~\cite{kirichenko2023last}) assumes access to \emph{group labels} that explicitly identify whether a sample contains a confounder, enabling reweighting or robust optimization strategies. Finally, a third set of approaches leverages Explainable AI (XAI) techniques to identify and prune spurious features directly (e.g.\ P-ClarC~\cite{anders2022finding}, RR-ClarC~\cite{dreyer2024hope, pahde2025patclarc}, Explanation Pruning~\cite{linhardt2024preemptively}). However, while promising in principle, XAI-driven methods have not yet fully delivered their potential. Attribution methods can miss important forms of spurious behavior, such as sensitivity to geometric transformations (e.g.\ rescaling, translation) or to color profile, and they are also unable to highlight the absence of a feature.

\medskip

\begin{figure*}[t]
  \centering
  \makebox[\textwidth][c]{\includegraphics[width=1.05\textwidth]{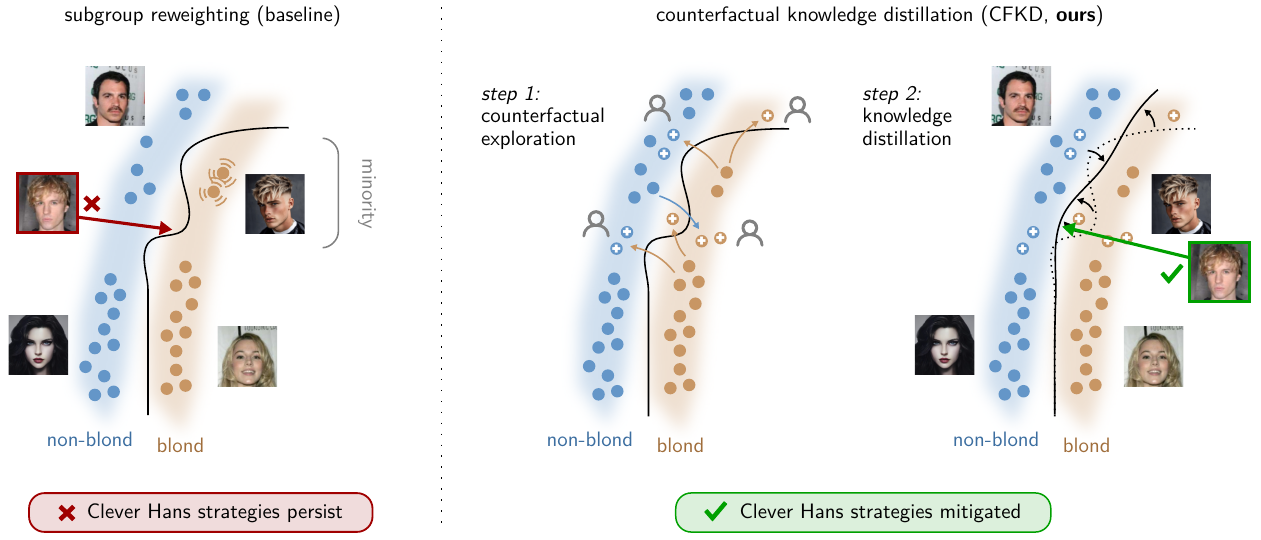}}
  \caption{
  Proposed CFKD method compared to a subgroup reweighting baseline (e.g.\ DFR). In the shown `blond' vs.\ `non-blond' classification task, the attribute `male/female' becomes a spurious correlate due to the low representation of `blond males' in the data, leading to a Clever Hans effect. CFKD addresses this by populating underrepresented groups through counterfactual generation. Specifically, generating counterfactuals for `blond females' occasionally results in `blond males', revealing the CH effect. CFKD then relabels these misclassified instances and adds them to the dataset (step 1) followed by retraining (step 2). This guided data augmentation/labeling/retraining strategy allows CFKD to address CH effects more directly than subgroup reweighting.
  }
  \label{fig:intro}
\end{figure*}

In this work, we extend \emph{Counterfactual Knowledge Distillation} (CFKD)~\cite{bender2023towards}, a novel approach that builds on the strengths of Explainable AI while avoiding the limitations of previous methods (cf.\ Figure~\ref{fig:intro}). CFKD presents \emph{factual} and \emph{counterfactual} examples to a teacher, who is then asked to determine whether the features that provides class evidence have been removed or persist. This process requires only domain knowledge (but no data science expertise), making it reproducible and broadly applicable. In doing so, CFKD provides a scalable pathway toward mitigating CH effects in a wide range of ML models, thereby contributing to the development of more robust, generalizable, and trustworthy AI systems. Importantly, CFKD does not require prior knowledge of specific confounders---a requirement that is often unrealistic in complex application domains, as it demands both technical expertise and deep domain insight.---Moreover, our algorithm is designed for use under Good Laboratory Practice (GLP) guidelines as a standalone application with minimal human supervision, where system validation requires decision-making strategies to be as transparent as possible. Our contributions are:

\medskip

\urlstyle{rm}

\begin{enumerate}[label=(\roman*)]
    \item \textbf{A novel framework} (CFKD) for eliminating spurious correlations via counterfactual explanations and teacher feedback.
    \item \textbf{A methodology} to benchmark and compare different model correction techniques.
    \item \textbf{A new histopathology dataset} ({\color{Blue}\url{https://huggingface.co/datasets/janphhe/follicles_true_features}}) with causal and spurious feature masks, designed for benchmarking explainable AI and robust ML.
\end{enumerate}

\medskip

\noindent We show that CFKD outperforms alternative approaches without requiring prior knowledge of the confounders on five different datasets, on different strengths of spurious correlations (even when the source of spurious correlation is unknown), on different dataset sizes, and when using a foundation model as a base or training from scratch. Moreover, we demonstrate the ability of different teachers to distill knowledge into students with the help of CFKD. Code for reproducing our results alongside implementations of different counterfactual explainers and group robustness algorithms is available at {\color{Blue}\url{https://github.com/Explainable-AI-Berlin/pytorch_explain_and_adapt_library}}.

\section{Related Work}

In this section, we present the four directions of research that are closest in terms of their goal to our method.
Given our experiment design we emphasize research specific to Clever Hans removal and focus less on research combating general overfitting.

\subsection{Data Augmentation} 
Data augmentation techniques have been widely explored to improve generalization and robustness of deep learning models. MixUp~\cite{zhang2017mixup} blends pairs of training examples and their labels to encourage smoother decision boundaries. CutOut~\cite{devries2017cutout} improves robustness by masking out random patches in the input, while CutMix~\cite{yun2019cutmix} replaces patches with regions from other images, combining both labels accordingly. RandAugment~\cite{cubuk2020randaugment} and AutoAugment~\cite{cubuk2019autoaugment} automate the search for optimal augmentation policies, and AugMix~\cite{hendrycks2020augmix} mixes augmentations to improve robustness and uncertainty calibration. Other works explore adversarial augmentations~\cite{volpi2018generalizing} and style transfer–based augmentations~\cite{geirhos2019imagenettrained} to enhance invariance.
Another way is to use diffusion models to augment the inputs (DiffAug) for better generalization~\cite{trabucco2023effective}.
We implement DiffAug with the same unconditional diffusion model that we also use for our counterfactual explainer.

In contrast to these works, which apply augmentations independently of the model's weaknesses, our approach uses augmentation steered by the model, specifically through counterfactuals that are informative about confounding features.

\subsection{Group Distributional Robustness against Spurious Correlations}
Several approaches aim to robustify models against spurious correlations using group information.
Group-DRO~\cite{sagawa2019distributionally} optimizes performance for the weakest group, but requires group labels for all data and costly retraining.
Just Train Twice (JTT)~\cite{liu2021just} relaxes this by only requiring group annotations on a small validation set, while Deep Feature Reweighting (DFR)~\cite{kirichenko2023last} avoids full retraining by reweighting the last layer. In our experiments, this family of approaches is represented through GroupDRO and DFR. GroupDRO is the seminal work in group distributional robustness, and DFR is a more recent, very simple, and yet very effective baseline.
However, methods like DFR have three fundamental limitations: 1) It is unclear how to attain group labels in the first place, and 2) even if one has group labels and upweights the minority groups, they do not cover the distribution inside the groups well due the low sample size and 3) the scaling to more than one spurious correlation is very bad, because groups become even smaller, making it even harder to detect the confounding directions.
CFKD, on the other hand, offers the following advantages: 1) a simple and constructive way to implicitly attain group labels without knowledge what the confounders are, 2) by sparse, reflective counterfactuals CFKD creates a distribution in the minority groups as rich as the majority groups only with the confounding feature flipped and 3) by diverse counterfactuals and implicit transfer of knowledge about the confounders, scaling to multiple confounders is no problem.

\subsection{XAI-Based Model Improvement}
Attribution-based methods such as Explanatory Interactive Learning (XIL)~\cite{schramowski2020making,teso2019explanatory} and Right for the Right Reasons (RRR)~\cite{ross2017right} use human feedback or attribution-guided losses to encourage correct model reasoning.
Class Artifact Compensation (ClarC)~\cite{anders2022finding, dreyer2024hope, bareeva2024reactive, pahde2025patclarc} extends this by explicitly detecting confounders (via SpRAy~\cite{lapuschkin2019unmasking}) and removing or introducing them through augmentative (A-ClarC), projective (P-ClarC), or Right-Reason (RR-ClarC) strategies.
Our method, Counterfactual Knowledge Distillation (CFKD), is related to ClarC but leverages counterfactuals both for confounder detection and augmentation, providing an integrated mechanism without relying on attribution maps.
This class of approaches is represented in our experiments by P-ClarC and RR-ClarC. P-ClarC is the seminal work for removing Clever Hans in image classifiers, and RR-ClarC is, to the best of our knowledge, the current state of the art.

\subsection{Counterfactuals for Dataset-Level Improvements}
Counterfactual data augmentation has been explored mostly in NLP, often with manually created counterfactuals~\cite{kaushik2019learning,deng2023counterfactual} or automated tools like Polyjuice~\cite{balashankar2023improving}, which remain dataset-level and invariant to the improved classifier. Active learning approaches also exist~\cite{margatina2021active}. Our work differs by leveraging counterfactuals that are tailored to the current model and embedded in an interactive feedback loop, ensuring that augmentations directly target spurious correlations.
Since all of these methods are intended for NLP instead of image classification, are applied on a dataset level only, and do not generate counterfactuals fully automatically, we do not compare against this class of approaches.

\section{Methods}

In this section, we present our main contribution, counterfactual knowledge distillation (CFKD), which enables us to refine a model (from now on referred to as the student) based on counterfactuals annotated through teacher feedback.
We continue introducing different methods for creating counterfactuals based on generative models. Finally, we present different teachers that can be used for CFKD and an estimator of the refined model's generalization capability, `feedback accuracy', which serves as a model selection criterion.

\subsection{Counterfactual Knowledge Distillation (CFKD)}

We propose the so-called Counterfactual Knowledge Distillation algorithm, which aims to distill knowledge from a teacher to a student and, as a result, to improve the generalization performance, e.g.\ in the presence of confounding variables. CFKD (illustrated in Fig.\ \ref{fig:intro} and detailed in Algorithm~\ref{alg:cfkd}) assumes (i) a trained classifier, (ii) a counterfactual explainer, (iii) a dataset, and (iv) a teacher. CFKD iterates over the dataset multiple times and proceeds at each iteration as follows: First, counterfactual samples are generated by the counterfactual explainer. Then, the teacher determines whether the counterfactual has changed the causal feature (True counterfactual) or not (False counterfactual). If it is a true counterfactual, the target label of the counterfactual search is selected. If not, the original label for the data point is selected. Then, the counterfactual is added to the training data with the appropriate label. The counterfactual image and selected label are then added to the training data. The classifier is retrained on the augmented training data, and the feedback accuracy (see Section \ref{sec:fa}) is measured. The algorithm continues for a specified number of iterations, and the refined model is selected based on the feedback accuracy. Contrary to many other knowledge distillation approaches, feedback can be provided by various sources of knowledge (including humans), since the image-counterfactual pairs are human-interpretable.

\begin{algorithm}[h]
\SetAlgoLined
\KwIn{Trained classifier $f$, dataset $D$, teacher $t$, number of iterations $n$}
\KwOut{Finetuned classifier $f'$}
\For{$i = 1$ to $n$}{
\For{$x, y \in D$}{
Choose $y_\text{target} \neq y$

Generate counterfactual image $\widetilde{x}$ based on $x$ and $y_\text{target}$

$t(x, \widetilde{x})$ decides if $\widetilde{x}$ is True or False counterfactual

\If{False counterfactual}{
Add the counterfactual image $\widetilde{x}$ and its true label $y$ to the training data
}
}
Retrain the classifier $f$ on the augmented training data

Measure the feedback accuracy
}
\KwRet{$f$}
\caption{Counterfactual Knowledge Distillation (CFKD)}
\label{alg:cfkd}
\end{algorithm}

\subsection{Counterfactual Explainers}

A variety of approaches have been proposed for producing counterfactuals of image classifiers, each aiming to overcome various challenges with the optimization problem itself and with the quality of the resulting counterfactuals. Diffeomorphic Counterfactuals (DiffeoCF)~\cite{ dombrowski2022diffeomorphic} and DiVE~\cite{rodriguez2021beyond} aim to produce counterfactuals that remain on the data manifold by using a generative model, specifically an invertible latent space model. The Adversarial Visual Counterfactual Explanations (ACE)~\cite{jeanneret2023adversarial} method utilizes gradients filtered through the denoising process of a diffusion model~\cite{ho2020denoising} to prevent moving in adversarial directions. It also uses a combination of L1 and L2 regularization between factual and counterfactual and RePaint~\cite{lugmayr2022repaint} as post-processing to keep pixel-wise changes to a minimum. Unlike ACE, which uses one noise level and re-encodes the current counterfactual at each iteration, DVCEs~\cite{augustin2022diffusion, augustin2024dig}, DiME~\cite{jeanneret2022diffusion}, and FastDiME~\cite{weng2025fast} encode only once in the beginning to a latent state that is then updated based on classifier guidance~\cite{dhariwal2021diffusion}. Ha et al.~\cite{ha2025diffusion} use a diffusion autoencoder to create counterfactual explanations. SCE~\cite{bender2025towards} builds upon ACE, removing weaknesses like non-convergence and adversarial counterfactuals while improving unbiasedness, sparsity, and diversity of the counterfactuals. As part of CFKD, we use the SCE approach, which avoids failure modes of other approaches, such as swapping the majority groups with each other and not changing the group distribution.

\subsection{Teacher Types}
\label{section:teachertypes}

To enable model improvement, we propose to connect CFKD with a variety of teacher types: 
`\textit{Human teachers}' constitute the most direct source of domain knowledge. To provide feedback, they require a user interface that can display data/counterfactual pairs and provide the option for the user to either discard or incorporate these counterfactuals, based on the user's assessment of their validity. Human teachers can provide feedback in a broad range of settings, however, the approach becomes impractical when conducting experiments on a large scale (e.g.\ requiring feedback on thousands of images) or when aiming for a fully reproducible evaluation pipeline.

`\textit{Oracle teachers}' provide a more reproducible alternative for tasks where privileged information is available, such as additional data without spurious correlations, and where a `ground-truth' ML model can be obtained from such data. The oracle teacher, equipped with this ground-truth ML model, can then deliver feedback based on the discrepancy between its own predictions (assumed correct) and the counterfactuals produced by the student model. Formally, denoting by $f_S$ the student network, by $f_O$ the oracle network, and by $x$ the original data point, we set the oracle teacher to discard counterfactuals $\widetilde{x}$ if they already fulfill $f_O(\widetilde{x}) = f_S(\widetilde{x})$, and to add them to the dataset otherwise. `\textit{Counterfactual analyzers}' are another type of artificial teacher, which instead of relying on an oracle ML model, inspect the counterfactual for properties such as spatial consistency, e.g.\ counterfactuals differing from the original image by changing meaningful image segments. These teachers are also endowed with privileged information, such as ground-truth segmentation masks.

\subsection{Model Selection Criterion}
\label{sec:fa}

In practice, it is also necessary to define a criterion for when to stop CFKD, or at which iteration CFKD has produced the best student model. The problem is non-trivial, because when the training and validation datasets are derived from the same pool of data, the validation data may be affected by the same spurious correlations as the training data, causing validation accuracy to be a poor indicator of true accuracy. To address this issue, we propose the feedback accuracy $\mathrm{ACC}_\mathrm{feedback} = N_\mathrm{correct} / N_\mathrm{total}$, measuring the proportion of counterfactuals generated on the validation dataset that are accepted as legitimate by the teacher. In essence, $\mathrm{ACC}_\mathrm{feedback}$ tracks the teacher's evaluation of the student model's ability to generate counterfactuals that correctly identify the generalizing strategy. In other words, a high feedback accuracy indicates that CFKD has produced a well-generalizing student model that is likely to perform accurately on the true data-generating distribution.

\section{Experiments}

In our experiments, we test our CFKD method in a broad range of contexts, ranging from fully synthetic data to a real-world follicle classification task. Our evaluation allows us to analyze the behavior of CFKD and SOTA methods in various settings, such as different levels of spurious correlation, different sample sizes, and different base models as starting points for knowledge distillation.

\subsection{Datasets}

We evaluate our method on five datasets, ranging from synthetic to real-world. The `\textit{Square dataset}' \cite{bender2025towards} consists of a simple 4-dimensional manifold embedded in image space (cf.\ Fig.\ \ref{fig:cfkd_before_after_qualitative} top row for examples). The four latent dimensions are the intensity of the foreground square, the intensity of the background, and the x- and y-position of the foreground square. A spurious correlation is introduced between the (relevant) foreground square's intensity and the (irrelevant) background square's intensity. This spurious correlation, combined with the background's higher saliency (high pixel footprint), causes the model to learn a Clever Hans strategy based on the background (cf.\ \cite{hermann2023foundations} for a related study). We also construct two synthetic datasets based on CelebA~\cite{liu2015faceattributes}: one which we call `\textit{CelebA Smiling vs.\ Copyright Tag}' where, randomly,
a copyright tag is pasted in the bottom right corner of each image and where the copyright tag's opacity is correlated to the Smiling attribute to induce a spurious correlation. A second variant, used in various papers~\cite{sagawa2019distributionally, liu2021just, kirichenko2023last} for benchmarking purposes, which we refer to as `\textit{CelebA Blond Hair vs. Male}' starts from an existing spurious correlation in the regular CelebA between the `blond' and `male' attributes. The spurious correlation can then be artificially strenghtened or weakened through subsampling. The `\textit{Camelyon17 dataset}'~\cite{bandi2018detection} is a histopathological dataset where the task is to determine whether a given patch of a histopathological slide contains cancer cells in its center. Camelyon17 is also annotated with metadata about the hospital from which the slides are coming. It is known that the hospital scanner and technical artifacts like staining, are a powerful confounder that occurs in the real world~\cite{komen2025towards}. We subsample different datasets from Camelyon17 to investigate the properties in question. These subsampled probes represent a realistic causal feature with a realistic confounder and a realistic spurious correlation.

\begin{figure*}[h]
  \centering
  \includegraphics[width=.9\textwidth]{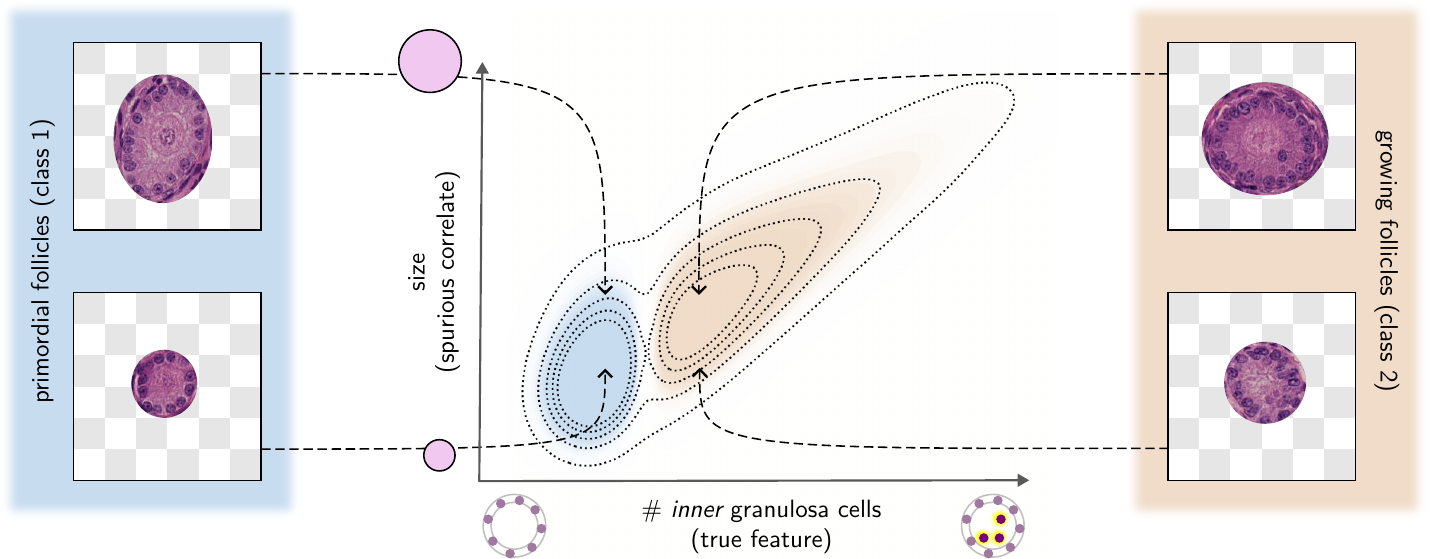}
  \caption{
  Cartoon depiction of the Follicle dataset and its spurious correlation. If the number of inner Granulosa cells is greater than or equal to one, a Follicle is considered to be of class 2 (`growing follicle'); otherwise, it is considered to be of class 1 (`primordial follicle'). The size of the follicle is, from a biological perspective, not determining the class, yet it is spuriously correlated to the number of Granulosa cells. Specifically, the dataset contains 1507 samples, with 776 small primordial (51.5\% of the data), 459 big growing (30.5\%), 155 big primordial (10.3\%), and 117 small growing (7.8\%).
  }
  \label{fig:follicle_dataset_intro}
\end{figure*}

We also consider a real-world use case that aims to test the toxicity of specific chemicals regarding the genesis of oocytes in rats' female offspring ovaries \cite{oecd443}. The test involves comparing the number of follicles of various types (e.g.\ `primordial and `growing') in groups of female offspring of treated and untreated rats, with follicle counts collected from histopathological images. Accurately counting follicles of each type is a time-consuming task, which appeals a ML approach. To this end, we assembled in collaboration with our industry partners a new histopathological dataset, which we call the `\textit{Follicle dataset}' and where the task is to classify whether each image contains a `primordial follicle' or `growing follicle' at its center. Biologically, the distinction between primordial and growing follicles is based on the architecture of the granulosa cell layers: primordial follicles are characterized by a single flattened layer of granulosa cells surrounding the oocyte, whereas growing follicles---encompassing both primary and secondary stages---exhibit at least one additional layer of cuboidal granulosa cells, thereby forming two or more concentric layers. A cartoon depiction of the dataset is given in Figure \ref{fig:follicle_dataset_intro}. During preliminary analyses, we identified a significant confounding factor within the dataset. Specifically, follicle size is strongly correlated with follicular stage: growing follicles are, on average, substantially larger than primordial follicles. Consequently, even simple size-based classification already yields high predictive accuracy, independent of the biologically relevant morphological features. For the model to implement the correct decision boundary, it must resist the tendency to base its decision on follicle size and solve instead the more complex task of counting the number of inner Granulosa cells. We regard the dataset as a realistic representation of histopathological conditions, since size differences naturally accompany follicular development in vivo. For our experimental design, the dataset was used as the reference standard, and parameters such as dataset size and the level of spurious correlation were modeled accordingly.

\subsection{Teachers Feedback}
\label{section:teachers}

Most of our experiments are based on an `\textit{oracle teacher}' where the auxiliary oracle network is trained on an extended dataset $\mathcal{D}^*$ where the artificial spurious correlation structure has been removed in order to more faithfully match the true underlying data distribution. For specific experiments (Fig.\ \ref{fig:teacher_comparison}), we also include a `\textit{human teacher}' who is given the opportunity to deliver feedback via a simple user interface displaying factual/counterfactual image pairs $(x,\widetilde{x})$ alongside two options for the human expert to discard or incorporate the counterfactual. As an additional teacher, we also include a `\textit{counterfactual analyzer}' (cf.\ Section \ref{section:teachertypes}), where the counterfactual difference $(x-\widetilde{x})$ is juxtaposed with a binary mask $M \in \{-1,1\}^d$ indicating where causal features are located in the image. Generated counterfactuals are then discarded or incorporated based on the sign of the dot product $\langle |x-\widetilde{x}|,M\rangle$. As a control, we also include a `\textit{random teacher}' that randomly discards or incorporates counterfactuals with equal probability. This baseline helps verify that it is the actual teacher feedback (and not a pure effect of data augmentation) that is the cause for model improvement.

\subsection{Results}

In this section, we report the performance of different model improvement techniques including our proposed CFKD, on various datasets, where we can furthermore control sample size, spurious correlation level, and other parameters such as the choice of teacher and underlying counterfactual generator. We measure the performance using the average group accuracy, which provides a balanced evaluation in the presence of group imbalance. Formally, we can always partition our data in four groups: (i) class $\omega_1$ with the confounder, (ii) class $\omega_1$ without the confounder, (iii) class $\omega_2$ with the confounder, and (iv) class $\omega_2$ without the confounder, which we denote by $\omega_1^+, \omega_1^-, \omega_2^+, \omega_2^-$ respectively. The average group accuracy (AGA) is then defined as:
\begin{align*}
\text{AGA} &= \Big.\text{avg}\,\big\{ \text{Acc}(\omega_1^+), \text{Acc}(\omega_1^-), \text{Acc}(\omega_2^+), \text{Acc}(\omega_2^-) \big\}\Big.
\end{align*}
Our main experiment consists of evaluating CFKD performance compared to other model improvement methods on the five datasets presented above. Here, we fix for the first four datasets the correlation parameter to $0.96$ and we also fix the number of instances available for model improvement to $1000$. Results are shown in Table~\ref{tab:sota_comparison}. We can observe that CFKD ranks first by a wide margin on all datasets except one, the Camelyon17 dataset, where CFKD is narrowly outperformed by RR-ClArC. It is important to note that CFKD was at a disadvantage in this setting because all baseline methods, including the XAI-based P-ClArC and RR-ClArC, were given perfect group labels. Overall, these results demonstrate that CFKD achieves strong model improvement in practice, clearly outperforming existing approaches.

\begin{table}[h!]
    \caption{
    AGA scores obtained by CFKD and other model improvement techniques on each dataset. For the Square dataset,  CelebA Smiling vs Copyrighttag, Blond and Camelyon17, the correlation parameter is set to 0.96, and 1000 samples were used. Best results are shown in bold, and second best with an underline.
    }
    \centering
    \small
    \begin{tabular}{lccccc}
        \toprule
        Method                      & Square         & Smiling        & Blond          & Camelyon     & Follicles \\
        \midrule
        Original                    & 51.1 & 51.3 & 72.7 & 55.3 & 65.9\\
        DiffAug                     & 54.9 & 63.1 & 72.7 & 55.3 & 65.9\\
        GroupDRO                    & 61.3 & 56.3 & 73.0 & 72.6 & 79.2\\
        DFR                         & 52.1 & 53.0 & \underline{77.5} & 71.9 & 73.3\\
        P-ClArC                     & 78.6 & 65.9 & 74.4 & 72.1 & 76.6\\
        RR-ClArC                    & \underline{78.6} & \underline{68.7} & 74.5 & \textbf{81.3} & \underline{79.8} \\
        CFKD (ours)                 & \textbf{94.5} & \textbf{79.6} & \textbf{79.1} & \underline{78.7} & \textbf{83.9} \\
        \bottomrule
    \end{tabular}
    \label{tab:sota_comparison}
\end{table}

\paragraph{Effect of Dataset Characteristics}

\begin{figure*}[h!]
  \centering
  \includegraphics[scale=.64]{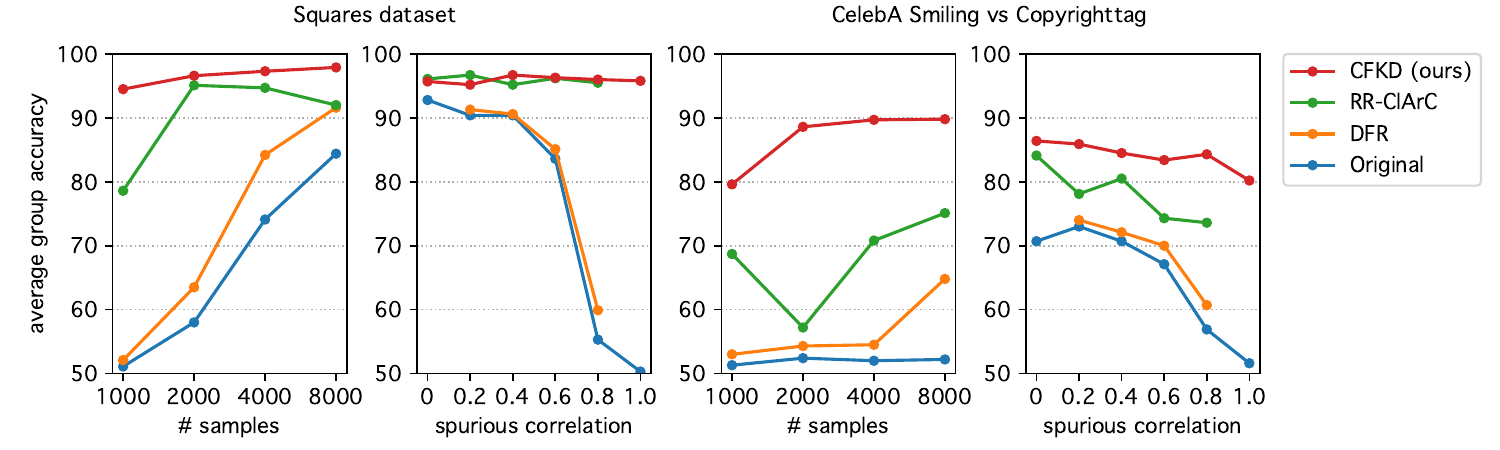}
  \caption{Effect of sample size and spurious correlation level on accuracy of benchmarked methods. Results are shown for the Square and the CelebA Smiling vs.\ Copyrighttag datasets. When analyzing the effect of sample size, we fix the correlation parameter to $0.96$. When analyzing the effect of correlation strength, we fix the sample size to $1000$. DFR and RR-ClArC results are not reported for a correlation parameter of $1.0$ as the methods are not applicable in that setting. For a correlation parameter of $0.0$, DFR reduces to the original training.
  }
    \label{fig:effect}
\end{figure*}

Figure \ref{fig:effect} analyzes the performance of the benchmarked methods under varying sample sizes and spurious correlation levels. With respect to sample size, one can see that CFKD already performs well with $1000$ samples while continuing to scale with more samples, while the Original model, RR-ClArC, and DFR fail with a few samples and then improve faster with more samples while remaining below CFKD. With respect to correlation, one can see that CFKD remains relatively stable for Smiling and very stable for the Square dataset with increasing the correlation parameter, while the performance of the original model, DFR, and RR-ClArC deteriorates sharply. Interestingly, CFKD remains effective with \textit{full spurious correlation}, meaning the minority can be empty, and the algorithm still works, a setting that none of the existing methods are able to handle. CFKD also remains effective with \textit{no spurious correlation}, implying that the some of gains have to come from distilling knowledge about other, unknown confounders into the model. We note that RR-ClArC's achieves similar improvement in the no correlation setting, but at the cost of extensive hyperparameter search.

\paragraph{Effect of Machine Learning Setup} Here, we analyze the behavior of CFKD beyond classical ML models trained from scratch, to also include foundation models. We focus for this experiment on the Camelyon17 dataset, where we can leverage the recent UNI~\cite{chen2024towards} foundation model. Results comparing CFKD with the original model and the DFR baseline are shown in Table~\ref{tab:model_comparison}. We observe that CFKD performs best across all ML setups, with the highest performance difference recorded on the classical setting of a ML model trained from scratch.

\begin{table}[h!]
    \caption{Evaluation of model improvement techniques under various ML setups, in particular, a Resnet-18 trained from scratch, and linear readouts built on the foundation model UNI~\cite{chen2024towards} with and without fine-tuning (UNI-F and UNI-L respectively). Experiments are done with training on $1000$ samples and with correlation parameter $0.96$ on Camelyon17. Highest accuracy is shown in bold.}
    \label{tab:model_comparison}
    \centering \small
    \begin{tabular}{lccc}
        \toprule
        Method       & ResNet-18 & UNI-F  & UNI-L  \\
        \midrule
        Original     & 55.3 & 67.9 & 70.6 \\
        DFR          & 71.9 & 68.3 & 73.6 \\
        CFKD (ours)  & \textbf{78.7} & \textbf{70.6} & \textbf{75.2} \\
        \bottomrule
    \end{tabular}
\end{table}

\begin{figure*}[h]
\centering
\includegraphics[scale=.64]{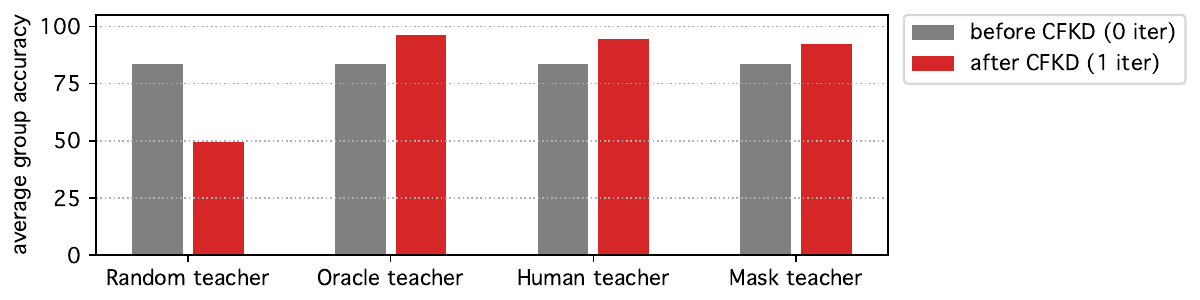}
\caption{A comparison between the different teachers that were used on the Square dataset with correlation parameter $0.6$. The performance is measured before and after applying one iteration of CFKD.}
\label{fig:teacher_comparison}
\end{figure*}

\paragraph{Effect of Teacher Type} In the following, we experiment with different teacher types described in Section \ref{section:teachers}. For this experiment, we focus on the Square dataset and set the spurious correlation parameter to $0.6$. Results are shown in Figure~\ref{fig:teacher_comparison}. We observe that the oracle, human, and mask teachers are all effective in improving the original model, with the oracle teacher achieving the best results, as expected. In comparison, the random teacher destroys the model completely with the confusingly labeled samples and degrades it back to random performance, showing that the benefit of teacher feedback arises from the injected domain knowledge and not mere data augmentation.

\paragraph{Effect of Counterfactual Explainer} As a last experiment, we conduct an ablation study, where we demonstrate the necessity for our method to rely on a strong counterfactual explainer. Specifically, we replace the advanced SCE counterfactual explainer used in our method by classical ones such as ACE, DiME and FastDIME. Results are shown in Table~\ref{tab:explainers}, and highlight the benefit of using SCE. This is due to various problems in other counterfactual explainers, already identified in \cite{bender2025towards}, such as being biased towards low pixel footprint features, not converging due to optimization issues, or lack of diversity. Each of these properties is important for CFKD.

\begin{table}[h!]
\caption{Ablation Study of the used counterfactual explainer on Square, Smiling, and Camelyon with 1000 samples and correlation parameter $0.96$. One can see that using SCE is essential for a stable performance gain.
}
\label{tab:explainers}
\centering \small
\begin{tabular}{lccc}
	\toprule
	Explainer  &  Square         & Smiling        & Camelyon \\
	\midrule
	\textit{Original}   &  51.1 & 51.4 & 55.3 \\
	ACE        &  51.1 & 79.4 & 55.3 \\
	DiME       &  51.1 & 62.1 & 48.4 \\
	FastDiME   &  88.1 & 60.6 & 50.8 \\
	SCE        &  \textbf{94.5} & \textbf{79.6} & \textbf{78.7} \\
	\bottomrule
\end{tabular}
\end{table}

\begin{figure*}[h!]
  \centering
  \makebox[\textwidth][c]{
  \rotatebox{90}{\sffamily \scriptsize ~~~~~~~~~~~~ Counterfactual ~~~~~~~~~~~~~~~~~ Original}~
  \rotatebox{90}{\sffamily \scriptsize  ~~~~ \em \color{gray!75!black} after CFKD ~~~~ before CFKD}
  \includegraphics[width=.95\textwidth,clip,trim=400 0 0 0]{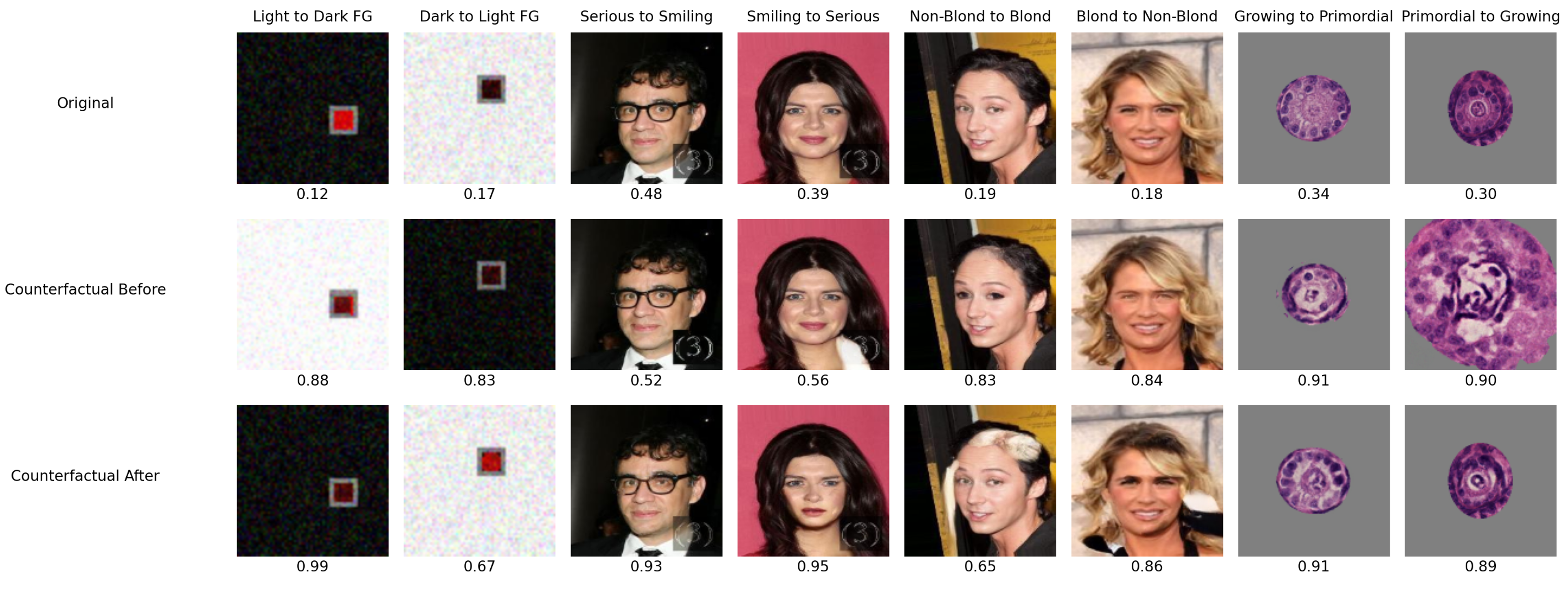}
  }
  \caption{
  Qualitative analysis of CFKD, where for a variety of data points from the datasets considered (top-row), we display one counterfactual generated before applying CFKD (middle row) and after applying CFKD (bottom row). These images highlight the transition from initially perturbing non-causal features (e.g.\ follicle size) to finally perturbing causal features (e.g.\ inner Granulosa cells).
  }
  \label{fig:cfkd_before_after_qualitative}
\end{figure*}

\paragraph{Qualitative Results} Lastly, we aim to verify qualitatively how the structure of the model is improved through the application of our CFKD method. We can obtain such insights by looking at SCE counterfactuals generated before and after applying CFKD model improvement (i.e.\ Original vs.\ CFKD in Table \ref{tab:sota_comparison}). These qualitative results are presented in Figure~\ref{fig:cfkd_before_after_qualitative}, where we show the original data point and the counterfactuals generated before and after model improvement. We can observe a clear evolution where initially, counterfactuals perturb the confounding features (and are thus new data points to be inserted in the distillation task), and where after applying CFKD, they all perturb the correct causal feature. In particular, for the Square dataset, one can see that before applying CFKD, the counterfactuals were built by changing the confounding background, and after they were built by changing the foreground. For Serious to Smiling and vice versa, counterfactuals were originally built by changing the confounding copyright tag, later on by changing the facial expression. For Non-Blond to Blond and vice-versa, counterfactuals first add/remove the confounding makeup, and only later change the hair color. For the Follicle dataset, we can see that the counterfactual corresponds initially to a change in the size of the follicle and only later on tamper with the causal inner Granulosa cells.

\section{Conclusion}

In this work, we have presented a novel technique, CFKD, to detect and remove reliance on confounders in deep learning models by generating counterfactuals and leveraging explanatory feedback. We demonstrated the effectiveness of CFKD at improving the model's accuracy on both synthetic and real-world datasets, especially on underrepresented subgroups. Unlike techniques based on subgroup reweighting, our method actively explores the original model's decision boundary to look for potential flaws, and correcting them. Our method is flexible and can work together with a variety of teacher models and counterfactual generation techniques.

So far, we have focused on applying and testing CFKD on image classification models. In practice, CFKD makes relatively few assumptions and should be extensible beyond image data. In future work, our CFKD approach could be extended to other data modalities. Application to multimodal data would be of specific interest as the high dimensionality could further amplify the characteristics of data we have studied in this paper, namely, the presence of near-perfect spurious correlations, and the increasing difficulty to identify data subgroups needed for applying reweighting techniques. We speculate that CFKD-like approaches may therefore be even more critically needed over existing model improvement methods.
Overall, our paper contributes to improving the interpretability and reliability of deep learning models and has the potential to impact a wide range of applications.

\section{Acknowledgements}

This work was supported by the German Ministry for Education and Research (BMBF) under Grant 01IS18037A, and by BASLEARN -- TU Berlin/BASF Joint Laboratory, co-financed by TU Berlin and BASF SE. KRM was funded in part by the Institute of Information \& Communications Technology Planning \& Evaluation (IITP) grants funded by the Korea Government (No.\ 2022-0-00984, Development of Artificial Intelligence Technology for Personalized Plug-and-Play Explanation and Verification of Explanation). We also want to thank Lorenz Linhardt, Marco Morik, and Jonas Dippel for fruitful discussions and helpful comments.

{
\bibliographystyle{elsarticle-num}
\bibliography{egbib}
}

\end{document}